\documentclass[acmtog, authorversion, nonacm]{acmart}

\AtBeginDocument{%
  \providecommand\BibTeX{{%
    \normalfont B\kern-0.5em{\scshape i\kern-0.25em b}\kern-0.8em\TeX}}}

\usepackage{mathtools}          % Extends amsmath, providing common math tools
\usepackage{mathrsfs}           % Enables \mathscr, which can work in cases that \mathcal does not
\usepackage{graphicx}           % For including images
\usepackage{subcaption}         % Allows for the use of subfigures and subcaptions
\usepackage[space]{grffile}     % For spaces in image names
\usepackage{url}                % For displaying urls

\usepackage{amsmath}
\usepackage{enumitem}
\usepackage{booktabs} % For better table rules
\usepackage{threeparttable} % For notes below table
\usepackage{float}

\usepackage[utf8]{inputenc}
\usepackage[linesnumbered,ruled,vlined]{algorithm2e}

\DeclareMathOperator*{\argmax}{arg\,max}
\usepackage{hyperref}

\setcopyright{acmlicensed}
\copyrightyear{2024}
\acmYear{2024}
\acmDOI{XXXXXXX.XXXXXXX}
\acmConference[Preprint]{}{Under review}
\begin{document}

\title{Fairness Incentives in Response to Unfair Dynamic Pricing}

\author{Jesse Thibodeau}
\affiliation{%
  \institution{Mila - Quebec AI Institute, HEC Montréal}
  \streetaddress{6666 Rue Saint-Urbrain}
  \city{Montr\'eal}
  %\state{Qu\'ebec}
  \country{Québec, Canada}}
\email{jesse.thibodeau@mila.quebec}

\author{Hadi Nekoei}
\affiliation{%
  \institution{Mila - Quebec AI Institute, Université de Montréal}
  \city{Montréal}
  \country{Québec, Canada}}
\email{nekoeiha@mila.quebec}

\author{Afaf Taïk}
\affiliation{%
  \institution{Mila - Quebec AI Institute, Université de Montréal}
  \city{Montréal}
  \country{Québec, Canada}}
\email{afaf.taik@mila.quebec}

\author{Janarthanan Rajendran}
\affiliation{%
  \institution{Dalhousie University}
  \city{Halifax}
  \country{Nova Scotia, Canada}}
\email{janarthanan.rajendran@dal.ca}

\author{Golnoosh Farnadi}
\affiliation{%
  \institution{Mila - Quebec AI Institute, McGill University}
  \city{Montréal}
  \country{Québec, Canada}}
\email{farnadig@mila.quebec}

\renewcommand{\shortauthors}{Thibodeau, et al.}

\begin{abstract}
The use of dynamic pricing by profit-maximizing firms gives rise to demand fairness concerns, measured by discrepancies in consumer groups' demand responses to a given pricing strategy. Notably, dynamic pricing may result in buyer distributions unreflective of those of the underlying population, which can be problematic in markets where fair representation is socially desirable. To address this, policy makers might leverage tools such as taxation and subsidy to adapt policy mechanisms dependent upon their social objective. In this paper, we explore the potential for AI methods to assist such intervention strategies. To this end, we design a basic simulated economy, wherein we introduce a dynamic social planner (SP) to generate corporate taxation schedules geared to incentivizing firms towards adopting fair pricing behaviours, and to use the collected tax budget to subsidize consumption among underrepresented groups. To cover a range of possible policy scenarios, we formulate our social planner's learning problem as a multi-armed bandit, a contextual bandit and finally as a full reinforcement learning (RL) problem, evaluating welfare outcomes from each case. To alleviate the difficulty in retaining meaningful tax rates that apply to less frequently occurring brackets, we introduce $\texttt{FairReplayBuffer}$, which ensures that our RL agent samples experiences uniformly across a discretized fairness space. We find that, upon deploying a learned tax and redistribution policy, social welfare improves on that of the fairness-agnostic baseline, and approaches that of the analytically optimal fairness-aware baseline for the multi-armed and contextual bandit settings, and surpassing it by \(13.19\%\) in the full RL setting.
\end{abstract}






\maketitle

\section{Introduction}
\label{intro}
Firms equipped with modern compute power and vast consumer data logs may enjoy the economic benefit of engaging in the business practice of dynamic (or personalised) pricing. In doing so, profit-maximizing firms are able to charge potential customers, or customer segments, individualized prices based on estimates of their willingness-to-pay. From an efficiency standpoint, dynamic pricing has been shown to increase firm profitability and sales speeds~\citep{schlosser2018dynamic}. However, the welfare implications of the practice are somewhat unclear, and may occasionally yield socially undesirable outcomes as evidenced in the markets for insurance and housing, among others~\citep{zhu2023neoliberalization, betancourt2022dynamic}. For instance, while health insurers are known to make extensive use of dynamic pricing, survey data reveal that members of the hispanic population in the U.S. are on average roughly twice as unlikely to have healthcare coverage as members of the black population, who are in turn twice as unlikely as members of the white and asian populations~\citep{quickstats}. In addition, substantial price differences between consumers may lead to negative perceptions of fairness~\citep{lee2011perceived}, which can have their own disparate impact on market outlooks with respect to buyer participation, thereby propagating existing disparities. In housing, new constructions are often priced such that they are more appealing to privileged groups, thus contributing to growing wealth gaps and gentrification. Other sectors where buyer distributions should reflect those of the underlying population are the financial services and public transportation.
\\
\newline Dynamic pricing is generally employed by firms to assign prices to consumer segments such that the expected profit derived from each is maximized. Thus, a profit-maximizing firm engaging in dynamic pricing rarely bears any weight to its resulting buyer distribution, which, if not reflective of the underlying population, may be considered unfair. In this research, we address the problem of such imbalanced buyer distributions. Specifically, we examine the issue of dynamic pricing under \emph{demand fairness}~\citep{cohen2022price}, in markets where buyer distributions should be proportional to that of the underlying population. We begin by showing how profit-maximizing price allocations may fail to satisfy demand fairness. We then employ nonlinear programming (NLP) at the firm level to demonstrate how producers willing to consider fairness in their price allocations may do so with often minimal sacrifice to profitability. We call this case \emph{self-regulation}. Building on this, we acknowledge that in many cases, expecting a profit-maximizer to take it upon itself to consider fairness in their price allocations is unreasonable. Thus, we consider how a benevolent social planner (SP) might leverage available policy tools, notably taxation and subsidy, in order to dynamically incentivize market participation among underrepresented consumer groups, while penalizing unfair firm behaviour. To achieve this, we train an agent using reinforcement learning (RL) methods to generate financial incentives aimed at improving demand fairness, i.e. reducing the distributional gaps in buyer populations. In addition, we allow it to learn a simple redistribution scheme whereby consumers are subsidized based on their group membership. We find that social welfare can be improved using taxation aimed at incentivizing firms to adopt fairer behaviours. With the inclusion of subsidy, we find that firm profits and fairness outcomes both increase on average relative to our positive control benchmark.
\\
\newline Throughout our work, we make use of RL methods, which have a growing presence in the body of literature surrounding dynamic pricing and mechanism design. Specifically, we apply soft actor-critic to a variety of mechanism design tasks, from multi-armed bandits, to contextual bandits, and finally to a full RL problem. In addition, RL has been applied to several domains such as modelling markets~\citep{kastius2021dynamic}, efficient energy pricing~\citep{lu2018dynamic, kim2015dynamic} and insurance pricing~\citep{nieuwenhuis2019accounting}, where RL was shown to be useful in such scenarios where demand for a specific product is either unobserved or misunderstood to the point that it is difficult to uncover via traditional analytical methods. 

Furthermore, the rise in use cases for AI in industry indicates that RL methods will likely continue to gain popularity for applications to dynamic pricing. Finally, RL methods are scalable and therefore appropriate for the complexity of large consumer bases. Our research proactively addresses the social challenges arising from dynamic pricing, notably via the following contributions:

\begin{itemize}[noitemsep,topsep=0pt,parsep=4pt,partopsep=0pt]
    \item Introducing a new framework featuring three distinct policy mechanisms aimed at addressing demand fairness within single-product markets;
    \item Implementing various fairness incentive mechanisms through a range of economic policy variants, formalized via multi-armed and contextual bandits, as well as a full RL formulation;
    \item Presenting~\texttt{FairReplayBuffer}, a replay buffer for the soft actor-critic (SAC) algorithm, designed specifically for learning fairness taxation;
    \item Proposing a combined approach of subsidy and taxation to effectively address demand fairness issues and enhance social welfare;
    \item Conducting a comprehensive evaluation of our proposed framework through a simulation study involving firms each addressing two distinct consumer behaviours~\footnote{Our code will be made publicly available upon acceptance of this work.}.
\end{itemize}

\section{Related Work}
\label{related work}
Given the interdisciplinary nature of our work, we conduct a literature review consisting of topics from economics, specifically in the subfields of welfare economics and consumer choice theory, as well as a broad overview of fairness applications in sequential decision making tasks. 

\subsection{Fairness in Dynamic Pricing}
The study of welfare within the context of policy design began as a field of economics and gradually found applications in the fields of operations research \cite{gallego2019revenue} and computer science \cite{das2022individual}. Welfare theory covers a broad variety of subtopics, which include definitions of fairness and how these interact given an incumbent policy. For instance, a recent paper by~\citet{cohen2022price} formalizes multiple fairness definitions for dynamic pricing in terms of price, demand, consumer surplus, and no-purchase valuation, while proving analytically that no two of these are simultaneously satisfiable. Our analysis uses the demand fairness definition  \cite{cohen2022price, kallus2021fairness} as it most closely measures the disparate impact that dynamic pricing may have on buyer distributions, and which is well motivated by applications in education and healthcare \cite{kallus2021fairness}. In addition,~\citet{bertsimas2011price} define proportional fairness, a fairness criterion ensuring that the relative welfare improvement among one population subgroup exceed the corresponsing welfare loss among another. Proportional fairness is useful in illustrating the welfare tradeoff between population subgroups under different policies. While the aforementioned works are related through their use of constrained optimization and linear programming,~\citet{maestre2019reinforcement} use RL to impose fairness using Jain's index, which they treat as a measure of fairness in the price allocations between groups. While we gain inspiration for our consumer demand curves from theirs, they assume that firms will be self-regulating, while we introduce a benevolent SP to generate fairness incentives.

\subsection{Economics Foundations}
In this work, we explore dynamics within single-product markets. Consumer demand behaviours are expressed as purchase probabilities, a formulation commonly applied to evaluate how consumers might react to price fluctuations. The concept of nonlinear consumer preference was first examined in economic theory by~\citet{becker1962irrational} and later formalized in seminal work by~\citet{mcfadden1972conditional} and continues to prove foundational in consumer choice modelling~\citep{models2002progress}. In addition, we examine welfare through the lens of fairness, as did by~\citet{fleurbaey2008fairness}, who advocates for not only fair outcomes but also fairness in the processes that lead to such outcomes.

\subsection{Fairness in Sequential Decision Making}
Fairness in sequential decision-making is a critical concern as algorithms increasingly influence societal outcomes, in fields such as healthcare ~\cite{rajkomar2018ensuring}, loan approval \cite{hu2022achieving}, and recommender systems \cite{recsysfairseq}. Existing studies have established foundational approaches to fairness and highlighted challenges in ensuring equitable algorithmic decisions over time.~\citet{joseph2016fairness} introduce fairness constraints in bandit algorithms to prevent long-term disadvantages for individuals or groups. \citet{gillen2018online} tackle the implementation of fairness when fairness criteria are undefined, proposing a framework for online learning that adapts to evolving fairness metrics.~\citet{yin2024long} explore the delayed impacts of fairness-aware algorithms, revealing how short-term equitable decisions can lead to unfair outcomes in the long run.

Within this body of research, the closest application to our work is the AI Economist \cite{zheng2020ai}, where agents interact in a simulated economy by exchanging goods and services, while a SP aims to learn a taxation strategy that improves social welfare, defined as the product of equality and productivity. While their work is effective at showcasing emergent behaviours among consumer-workers under incumbent tax regimes, our focus lies on how an SP can impact societal outcomes by influencing firm objectives. In addition, while the AI Economist uses a fixed uniform tax redistribution policy, we formulate the subsidy as an additional component to the learning problem, as do~\citet{abebe2020subsidy}, who explore approaches for welfare-maximizing subsidy allocation under income shocks. However, while they use \textit{min-sum} and \textit{min-max} formulations, we consider average welfare across consumer groups.

\section{Problem Formulation and Methodology}
\label{formulations}
%\afaf{Add an introduction to this section :)}
In this section, we first define a simulated economy, outlining the consumer dynamics and possible scenarios in which a social planner might interact with firms. We follow with definitions for our firm and social planner learning problems, and finish with a discussion on the solution methods employed.

\subsection{Consumer Environment}
\label{sec:consumer_env}
\begin{figure*}[!ht]
\begin{center}
    \includegraphics[scale=0.285]{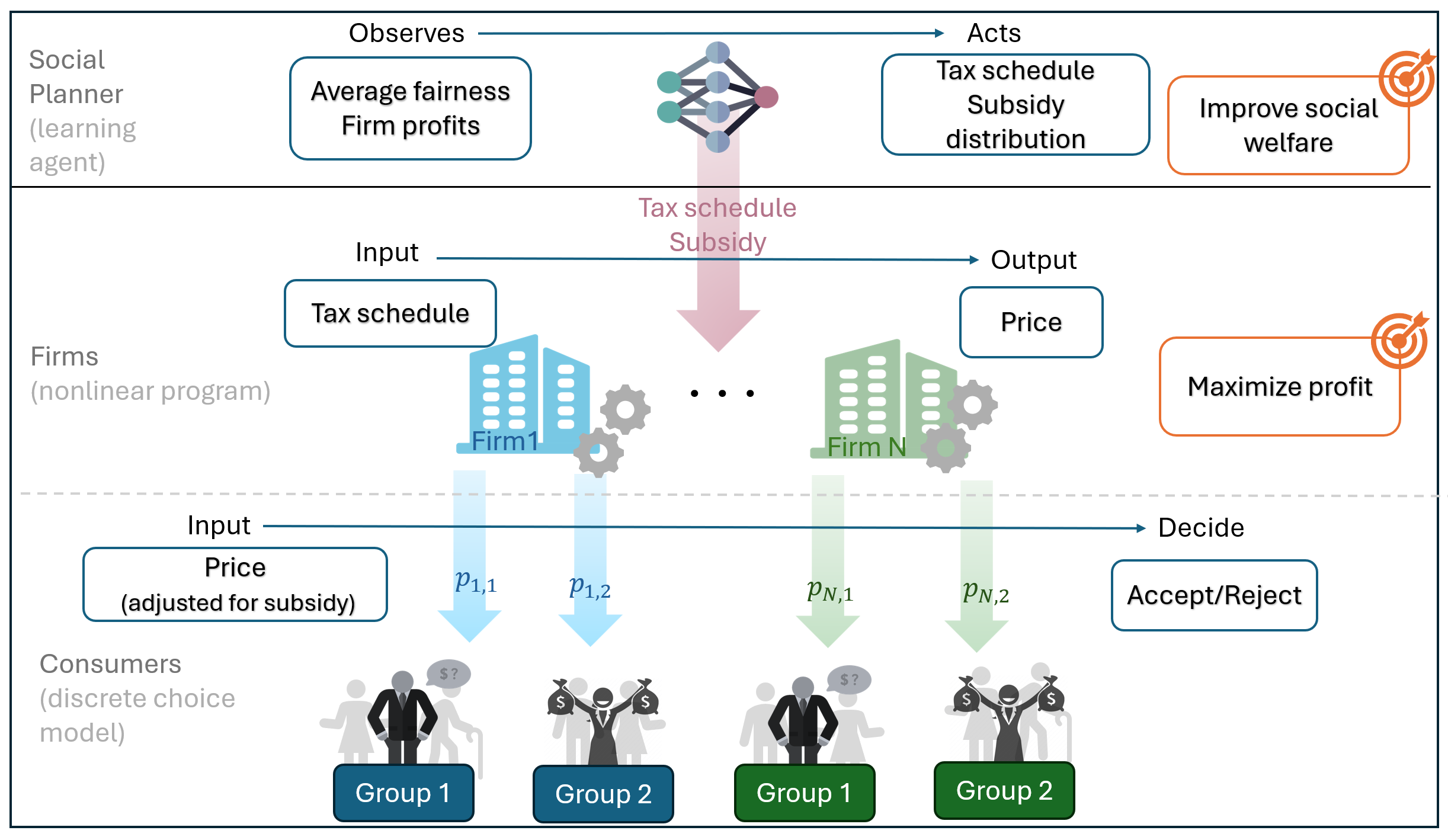}
\end{center}
\caption{Firms are capable of efficiently learning profit-maximizing pricing assignments from consumer demand responses. The social planner (SP) learns these implicitly through firms' fairness and profit scores and designs incentive mechanisms that use taxation and subsidy, pushing firms to minimize the gap in demand responses between consumer groups.}
\label{fig:2-level-system}
\end{figure*}

A firm's objective in using dynamic pricing is to charge the maximal price that a prospective consumer is likely willing to pay based on demand estimates. We design a single-product market consisting of distinct consumer groups, wherein members from each group decide whether to purchase a product at a price determined by the firm. Each consumer group $i$ has a unique purchase probability distribution, expressed as a discrete choice model by
\begin{equation}
    \mathbb{P}_{i}(\mathrm{purchase}=1\ |\ p) = [1 + e^{-(b_i + w_i \times p)}]^{-1},
    \label{eq:purchase_probability}
\end{equation}
where $p$ is the price assigned to the good, and parameters $b_i$ and $w_i$ capture different characteristics of consumer profile $i$ regarding their sensitivity to price fluctuations. Once a price assignment has been made by the firm, we obtain each consumer profile's purchase probability from~\autoref{eq:purchase_probability}. By linearity of expectation, we therefore anticipate that the number of consumers from group $i$ that purchase a product at its given price can be computed by $E[n_i] = N_{i}\times\mathbb{P}_i(\mathrm{purchase}=1)$, where $N_{i}$ is the number of consumers belonging to group $i$. In our experiments, we approximate these outcomes by simply passing these probabilities into a sequence of $N_i$ Bernoulli trials, such that $n_i \sim \mathcal{B}(N_i, \mathbb{P}_i(\mathrm{purchase}=1))$, where $n_i$ is a Bernoulli random variable corresponding to the resulting number of consumers from group $i$ who purchased the good. This provides an element of stochasticity to our environment, where purchase outcomes may vary over constant price assignments.
For simplicity, we consider the problem of a firm targeting two consumer groups, where one group has a higher sensibility to the price compared to the other one, which makes the consumers of this group underrepresented in the customer base of the firm. The fairness-agnostic firm might aim to maximize its profits with no regard for fairness, while its fairness-aware counterpart might weigh fairness outcomes into its maximization objective. In most real-world applications, we would not expect firms to explicitly self-regulate for fairness, as dynamic pricing is typically used to satisfy classical definitions of utility involving economic productivity (i.e., output, profit-maximization). To mitigate this, we introduce a dynamic social planner that aims to incentivize demand fairness by generating a tax schedule which determines the tax that will be levied on a firm based on their performance in terms of both profitability and fairness. In addition, we allow the social planner to redistribute the tax it collects to consumers as a subsidy to encourage higher market participation among underrepresented groups. Throughout our experiments, we consider demand fairness, and thus are concerned with the gap in purchase probabilities between consumer groups. We formalize our fairness notion as
\begin{equation}
    \mathrm{fairness}(p) = \ 1 - |\ \mathbb{P}_1(\mathrm{purchase} = 1\ |\ p) - \mathbb{P}_2(\mathrm{purchase} = 1\ |\ p)\ |\ ,
\label{eq:fairness}
\end{equation}
so that smaller gaps in demand correspond to higher fairness scores. Given this notion of fairness, the social planner's ultimate objective is to maximize social welfare, denoted by
\[
\texttt{swf} = \texttt{profit} \times \texttt{fairness},
\]
which illustrates the nonlinear tradeoff between firm productivity and fairness outcomes~\citep{zheng2020ai, bertsimas2012efficiency}. In~\autoref{fig:2-level-system}, we illustrate the economic process of the SP agent generating a tax and subsidy mechanism based on the current welfare context. Below it, a firm sets prices such as to maximize profits under the SP's generated mechanism. Finally, the environment responds, whereby consumers decide whether to accept or reject firms' price outputs.
\subsection{From Policy Preferences to RL Design}
In the real world, governments~\footnote{We use~\emph{government},~\emph{policy planner} and~\emph{social planner} interchangeably} might adopt varying policy plans dependent upon their maximization objective. In this light, our exploration aims to provide an overview of possible policy scenarios, discussing a range of formulations that an incentive mechanism can take on while associating each to different hypothetical use cases. These include (i) a %generalizable 
fixed policy mechanism, (ii) an adaptive policy mechanism conditioned on the current economic environment and (iii) an evolving policy framework that allows for frequent and ongoing changes to existing mechanisms. These are formulated as a multi-armed bandit problem, a contextual bandit problem and finally, a full RL problem. Below, we provide a detailed explanation of each policy mechanism.

\subsubsection{Fixed Policy Mechanism: Multi-armed Bandit problem}
\label{mab}

If a policy planner aims to deploy an incentive mechanism to regulate a new market with uncertain or poorly understood dynamics, they might consider designing a tax and subsidy framework based on simulated data or data from economies with analogous market structures. In this scenario, solving a multi-armed bandit problem would be the most appropriate approach, with policy actions tailored to suit the average firms represented in the data.

\subsubsection{Contextual Policy Mechanism: Contextual Multi-armed Bandit Problem}
\label{c-mab}

If the government was able to experiment with taxation frameworks that are conditioned on current market dynamics, it would want to create a dynamic policy mechanism. This can be formulated as a contextual bandit problem with an SP capable of generating incentives based on the observable economics context.

\subsubsection{Evolving Policy Mechanism: RL Problem}
\label{rl}
Let us further suppose that an SP is at will to make frequent changes to policy frameworks, under the assumption that firms are equally adaptive and will always compute their best response to incumbent policies. Treating this as a full RL problem, a policy planner may adapt taxation and subsidy to maximize their notion of social welfare over time.

\subsection{Firms}
% \afaf{Let's split this into 2 subsections, 3.3 Firms, 3.4 Social planner
% and 3.4 could include 3.2 actually and it would make more sense organization-wise}
The~\emph{firm} component of our problem formulation consists of (i) a nonlinear program (NLP) for the fairness-agnostic firm, and (ii) a modified NLP for the fairness-aware firm.

\subsubsection{Fairness-agnostic Firm}
Having defined a single-product market environment~(\autoref{sec:consumer_env}), we proceed to formulate the fairness-agnostic firm's problem as the nonlinear program:
\begin{equation}
\begin{aligned}
\max_{p} \quad & \sum_{i=1}^{G}n_{i}p\\
\textrm{s.t.} \quad & 0< p \leq p_{\mathit{max}} \\
\end{aligned}
\label{eq:fairness-agnostic max}
\end{equation}
where $G$ is the number of consumer groups in the market, and $n_{i}$ is the number of consumers from group $i$ who purchase and is itself a function of firm price $p$, as described in~\autoref{eq:purchase_probability}. \(p_{\mathit{max}}\) may be a value set by firms or governments representing some hypothetical maximum price that can be assigned to the product at hand. Thus, this is a straightforward revenue-maximization problem faced by the firm.

\subsubsection{Fairness-aware Firm}
In addition, we consider the case where a firm may seek endogenously to maximize some objective which takes the fairness outcomes of their price allocations into account. In similar fashion to~\autoref{eq:fairness-agnostic max}, this self-regulating firm's problem becomes:
\begin{equation}
\begin{aligned}
\max_{p} \quad & \sum_{i=1}^{G}n_{i}p \times \mathrm{fairness}\\
\textrm{s.t.} \quad & 0< p \leq p_{\mathit{max}} \\
\end{aligned}
\end{equation}
where the second term in the product enforces demand fairness by penalizing differences in demand between groups, as expressed in~\autoref{eq:fairness}.

\subsection{Benevolent Social Planner}\label{sec:benevolent_social_planner}
Given the range of possible preferences for policy makers discussed above, we continue with a mathematical formalism for our dynamic agents, including alternative formulations for our SP agent dependent upon possible policy objectives of government. We first introduce the shared formulations of the SP agent variants, then we outline their differences when introducing the sequential component to the full RL variant.

\subsubsection{SP problem formulation}
To alleviate the requirement for firms to proactively include fairness considerations in their pricing strategy, we introduce a benevolent SP to incentivize fair firm behaviour by leveraging the policy tools of taxation and subsidization. Specifically, the SP should generate a tax schedule which penalizes the firm based on its \textit{fairness bracket}, and redistributes wealth so as to narrow the distributional gap between market participants. In this scenario, we assume that the SP adopts a revenue-neutral policy; at the end of every tax period, its tax revenue is fully offset by its expenditure, which here takes the form of a consumption subsidy awarded to consumers in proportions determined by their group membership, thereby closing the cycle of wealth in our simulated economy. The SP is formulated as a learning agent that evolves in an environment containing $F$ firms, and each of the firms sets prices for a subset of the population constituted of individuals from both underrepresented and over-represented consumer groups.

\par For the cases where the SP is a contextual bandit or an RL agent, it observes a context vector from the context space \(\mathcal{X}^{SP} = \{f, \boldsymbol{\phi}\}\), where \(f \in [0,1]^{F}\) denotes a vector of individual firm fairness values, and $\boldsymbol{\phi} \in [0, p_{max}]^{F}$ denotes a vector of firm profits, normalized per customer, and takes actions from a continuous action space \[\mathcal{A} = \left\{ (A_1, A_2) \mid A_1 \in [0, 1]^{b}, A_2 \in [0, 1] \right\},\] where \( A_1 \) is a vector of size \(b\) with elements \( \tau \in [0, 1] \) denoting tax rates for each of \(b\) tax brackets, and \( A_2 \in [0, 1]\), denoting the proportion of the collected tax budget that will be awarded to the underrepresented group as a consumption subsidy in the following period. It follows that the remaining proportion of the tax budget \( 1 - A_2 \) will be distributed to the relatively overrepresented group. Therefore, to determine the effective price faced by a customer from the underrepresented group \(1\) and majority group \(2\), we update $p$ via
\begin{equation}
p^{t}_{1} := p^{t} - \frac{A_2 \times \mathcal{B}}{n_{1}} \quad ; \quad p^{t}_{2} := p^{t} - \frac{(1 - A_2) \times \mathcal{B}}{n_{2}},
\label{eq:price_update}
\end{equation}
where \(\mathcal{B} = \sum^F_{i=1} \tau_{i}^{t-1} \times \phi_{i}^{t-1}\) denotes the total tax revenue generated by \(F\) firms in the previous period, and \(n_{1}, n_{2}\) denoting the number of consumers belonging to group \(1\) and \(2\) respectively. 

\subsubsection{Social Welfare Over Time}

For the multi-armed and contextual bandit policy scenarios discussed in~\autoref{mab} and~\autoref{c-mab}, we formulate the social planner's problem as one of setting a tax schedule and redistribution scheme that will remain in effect over a time horizon \(T\). Rewards in these settings are therefore computed only at time \(T\). 
We further assume that the SP is able to compute fairness values by evaluating the buyer distribution resulting from the firm's learned pricing strategy. Consequently, the SP's reward is formulated as
\begin{equation}
    \mathcal{R}^{SP} = \frac{1}{F}\sum_{i=1}^{G_{j}}\sum_{j=1}^{F}n_{i}p_{j}(1-\tau_j) \times \mathrm{fairness}_{j},
\label{eq:SP_reward}
\end{equation}
where $F$ is the number of firms participating in the economy, and $G_{j}$ is the number of consumer groups addressed by firm $j$. We note that, with the addition of taxation, the firm's maximization objective becomes \(\max_{p}\sum_{i=1}^{G}n_{i}p \times (1-\tau)\) because they care about maximizing net profits.

For the RL solutions discussed in~\autoref{rl}, the SP's objective is to maximize discounted reward over the same horizon. In the RL setting, the SP agent's reward function is denoted
\[
\mathcal{R}^{SP}_{\text{RL}} = \sum_{t=1}^{T} \gamma^{t-1} \left( \frac{1}{F} \sum_{j=1}^{F} \sum_{i=1}^{G_{j}} n_{i,t} p_{j,t} (1-\tau_{j,t}) \times \mathrm{fairness}_{j,t} \right),
\]
where \(\gamma \in [0, 1]\) is a discount factor.
\subsection{Solution Methods}
In this section, we outline the methods employed. The SP agent learns its policy for choosing good incentive mechanisms given the welfare context using soft actor-critic (SAC), an RL algorithm designed to solve Markov decision processes, where there are temporal transition dynamics. S with emphasis on sample efficiency. SAC draws samples from a \emph{replay buffer} to perform policy gradient updates. Upon experimentation, our SAC agent expressed difficulty in generating taxation frameworks which were consistent in retaining meaningful information from lower brackets, which become less common as firms become fairer, leading us to introduce~\texttt{FairReplayBuffer}, which alleviates this concern.

\subsubsection{Soft Actor-Critic}
In our study, we employed the soft actor-critic (SAC) algorithm~\citep{softactorcritic}, an advanced RL technique suited for continuous action spaces. SAC is distinguished by its incorporation of entropy maximization into the reward function, promoting exploration while simultaneously striving for optimal policy development. This approach enhances the algorithm's sample efficiency and stability, making it particularly suited for environments where precise, adaptive control is required. While typically used for sequential decision-making over time, SAC is also appropriate for our bandit setup, as its sample efficiency makes it applicable to public policy where experimenting over long periods entails risk.

\subsubsection{FairReplayBuffer}
A fundamental challenge in generating meaningful incentives for less frequently %lesser 
seen contexts across the fairness space lies in how our SAC agent performs gradient updates from a traditional First-In-First-Out (FIFO) replay buffer, which would mostly consist of common firm behaviours in training. Due to this, a learning agent in our setting would gradually forget any information gained from infrequently observed regions of the observation space, thereby reducing its effectiveness on out-of-sample fairness brackets.
%Due to their infrequency, examples laying on the extremes of the observation space would not be impactful in the learning process, but for the sake of our learning agent's deliverable, which is meant to include some interpretability component, we designed a replay buffer that stores examples uniformly across the fairness and profit spaces.
%The protocol in~\autoref{alg:fair_buffer} describes the process by which we populate a \texttt{FairReplayBuffer}, which is crucial for our SAC agent to generate a deliverable which has some loosely monotonistic property between common examples, and their extreme neighbours. Ablations for~\texttt{FairReplayBuffer} may be found in~\autoref{ablation}.
To address this, we define a replay buffer that uniformly stores examples from across the fairness space, allowing the learning agent to output meaningful tax actions for each region. The procedure by which we populate a \texttt{FairReplayBuffer} is outlined in~\autoref{alg:fair_buffer}. The algorithm, along with ablations, can be found in~\autoref{fair_replay_buffer_appendix}.

\section{Experimental Design and Analytical Baselines}
\label{sec:experimental_design}
In this section, we define our working examples, which are designed to demonstrate the effects that a benevolent SP may have on market dynamics. We begin by defining four firms, each facing varying demand distributions between two consumer groups. Then, we record each firm's optimal pricing strategy and resulting fairness scores under profit-maximizing objectives, with and without self-regulation for fairness, in the absence of policy intervention. We set~\(\mathit{p_{max}} = 10\). Finally, we allow the SP to dynamically learn a taxation framework with respect to profit and fairness outcomes which will push firms to adopt prices which promote fairness levels approaching those achieved by fairness-aware firms.

\subsection{Multi-firm Setup}
\begin{figure*}[ht!]
    \centering
    \includegraphics[scale=0.45]{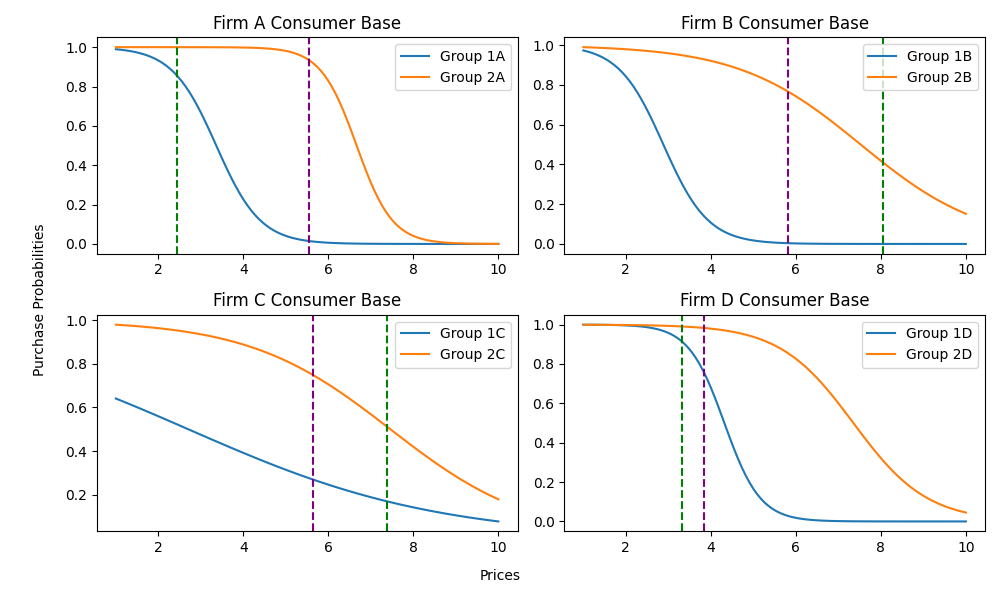}
    \caption{Consumer profiles: Each firm serves two consumer groups. For each, group 1 may be considered to have lower tolerance to rising prices than group 2. The vertical purple and green lines represent the analytical profit-maximizing price assigned by each firm in the fairness-agnostic and fairness-aware cases respectively, with the resulting vertical gaps between the orange and blue lines illustrating important discrepancies in purchase probabilities between consumer groups under price allocations associated to both behaviours.}
    \label{fig:consumer_profiles_multifirm}
\end{figure*}

We conduct our experiments in an environment where each of four firms addresses two distinct consumer groups. To describe their behaviours via purchase probability distributions, we apply parameters $b$ and $w$ found in~\autoref{appendix:consumer_distribution_params} (some of which are borrowed from~\citet{maestre2019reinforcement}), to~\autoref{eq:purchase_probability}, in order to obtain the curves depicted in~\autoref{fig:consumer_profiles_multifirm}. Using the ground-truth purchase probabilities embedded in our consumer environment, we can analytically derive optimal prices for each firm's objective. For instance, the fairness-agnostic firm's optimal price allocation is obtained by finding
\[\argmax_{p} \ \mathbb{P}_{i}(\mathrm{purchase}=1\ |\ p) \times p.\]

\subsection{Baselines}
Assuming firms are efficient at finding profit-maximizing prices under current market conditions (i.e., demand distributions and incentive schemes), and carrying our working example forward, we would expect fairness-agnostic firms to converge to the prices and fairness values in~\autoref{table:baseline_multifirm_profit} in the absence of policy intervention.
\begin{table}[ht]
    \centering
    \begin{subtable}{.45\textwidth}
        \centering
        \begin{tabular}{c|c c c c|c}
            \textbf{Firm\textsuperscript{agnostic}} & \textbf{A} & \textbf{B} & \textbf{C} & \textbf{D} & \textbf{Avg} \\
            % \hline
            % Price & 5.54 & 5.83 & 5.64 & 3.84 \\
            \hline
            \(f\) & 0.08 & 0.24 & 0.52 & 0.78 & 0.41 \\
            % \hline
            \(\phi\) & 2.63 & 2.24 & 2.87 & 3.34 & 2.77 \\
            \hline
            \(swf\) & 0.21 & 0.54 & 1.49 & 2.61 & \textbf{1.14} \\
        \end{tabular}
        \caption{Fairness, profit, and corresponding social welfare scores achieved by fairness-agnostic firms A, B, C and D in the absence of policy intervention.}
        \label{table:baseline_multifirm_profit}
    \end{subtable}\hfill % Use \hfill to add space between the subfigures
    \begin{subtable}{.45\textwidth}
        \centering
        \begin{tabular}{c|c c c c|c}
            \textbf{Firm\textsuperscript{aware}} & \textbf{A} & \textbf{B} & \textbf{C} & \textbf{D} & \textbf{Avg} \\
            % \hline
            % Price & 2.44 & 8.04 & 7.39 & 3.33 \\
            \hline
            \(f\) & 0.85 & 0.59 & 0.66 & 0.92 & 0.76 \\
            % \hline
            \(\phi\) & 2.27 & 1.65 & 2.51 & 3.17 & 2.40 \\
            \hline
            \(swf\) & 1.93 & 0.97 & 1.66 & 2.92 & \textbf{1.82} \\
        \end{tabular}
        \caption{Fairness, profit and corresponding social welfare scores achieved by fairness-aware firms A, B, C and D, who self-regulate to .}
        \label{table:baseline_multifirm_fairness-profit}
    \end{subtable}
    \caption{Resulting fairness (\(f\)), profit (\(\phi\)) and social welfare (\(swf\)) values from analytical optimal price assignments for fairness-agnostic and fairness-aware firms.}
\end{table}
By solving for the profit-fairness objective, also in the absence of policy intervention, i.e. self-regulation, we find that firms may change their prices significantly while sometimes achieving profits comparable to those in the fairness-agnostic case. These welfare-maximizing results for the fairness-aware firm~(\autoref{table:baseline_multifirm_fairness-profit}) serve as analytical baselines which can be used as a positive control for comparison with the results from our bandit and RL experiments. Further, referring back to~\autoref{fig:consumer_profiles_multifirm} provides an alternative visualization of fairness outcomes, where the difference between the orange and blue curves is the gap in demand between consumer profiles resulting from the firms' price assignments. We finally note from~\autoref{table:baseline_multifirm_profit} that our initializations for each firm lead them to converge to profit-maximizing prices resulting in a broad coverage of the fairness space. This is a design choice to demonstrate how optimal decisions from the firms' standpoint can have unforeseeable fairness implications.

\section{Empirical Evaluation}
We deploy a learning agent in the experimental setting defined in~\autoref{sec:experimental_design} using our multi-firm setup. While we initialize firms as described, we allow for a degree of stochastic fluctuations in consumer behaviour to occur, reflecting how demand can change subtley in the short-run. To reflect this, we sample purchase probabilities from a normal distribution around means given by the output of the sigmoid demand curves denoted in~\autoref{eq:purchase_probability}. The curves themselves are parameterized by \(w_{i}\) and \(b_{i}\), as found in~\autoref{appendix:consumer_distribution_params}. This indicates that the social planner may experience a different reward from having taken identical actions. This, paired with the Bernoulli trials which determine a consumer group's real purchase outcomes, %injects a level of noise in our environment which complexifies our learning problem. 
introduces a degree of noise into our environment, thereby adding complexity to our learning problem. The purchase probability for a consumer group in a given training step is therefore sampled as \(\mathbb{P}_{i,j} \sim \mathcal{N}(\mathbb{P}_{i,j}, \sigma^{2})\). In our experiments, we set \(\sigma=0.05\), allowing for small fluctuations in demand. We show the best results in terms of \(swf\) below. All experiments were run for 20 seeds. Our experiments are designed to answer the following questions:
\begin{enumerate}
    % \item Do dynamically generated incentive mechanisms improve social welfare?
    % \item How might a dynamic policy framework impact individual market participants?
    % \item What intuition can be drawn from the patterns exhibited by the learned mechanisms?
    \item What effects do different policy mechanisms, reflected through RL design choices, have on welfare? 
    % \item How do the incentive mechanisms introduced by the SP compare with an ideal self-regulating firm?
    \item How are individual market participants affected by a dynamic policy mechanism?
    \item How can RL-generated policies be adopted to assist policy makers and what are the long-term effects of such policy designs?
\end{enumerate}
We train the SP to tax firms based on where they lie along the fairness dimension, which we discretize into 5 brackets, using the multi-armed bandit, contextual bandit, and full RL formulations (see~\autoref{formulations}). Simultaneously, the SP chooses a proportion of the tax budget that it will re-distribute to the underrepresented consumer group in the next timestep. Our results show that, due to efficient wealth redistribution, the introduction of dynamic policies can approach and even improve upon the global welfare obtained from the analytical optimal case where firms are self-regulating.

\subsection{RQ1: Welfare Effects of Various RL Methods}
Here we evaluate the impact each formulation has on global welfare.~\autoref{table:aggregate_results} breaks down the average results for the multi-armed bandit, contextual bandit and full RL settings. 
First, we note that the RL SP yields the best \(swf\) results, on-par with the fairness-aware firm baseline, for which results are shown in \autoref{table:baseline_multifirm_fairness-profit}, illustrating the effectiveness of the proposed incentive mechanisms. 
However, while there is a clear pattern of improvement in \(swf\) as the problem formulation complexifies, it is worth noting the limitations of each implementation. While a multi-armed bandit SP underperforms relative to other solution methods, it nonetheless significantly improves welfare outcomes compared to profit-maximizing (fairness-agnostic) firms while requiring the least amount of data from the environment. This is suitable, under stationarity assumptions, for new markets with unknown dynamics.

%, later examining the tradeoffs which occur on an individual firm basis
\begin{table*}[h!]
  \centering
  \label{table:all_results}
  \small
  \begin{tabular}{c|c c c c c|c c c c c|c c c c c}
    \hline
    & \multicolumn{5}{c|}{Multi-armed bandit (\(S=0.63\))} & \multicolumn{5}{c|}{Contextual bandit (\(S=0.65\))} & \multicolumn{5}{c}{Full RL (\(S=0.66\))} \\
    \hline
    Firms & A & B & C & D & \textbf{Avg} & A & B & C & D & \textbf{Avg} & A & B & C & D & \textbf{Avg} \\
    \hline
    $f$ & 0.72 & 0.66 & 0.58 & 0.93 & 0.72\(\pm\)0.04 & 0.87 & 0.80 & 0.58 & 0.94 & 0.80\(\pm\)0.02 & 0.84 & 0.79 & 0.58 & 0.94 & 0.79\(\pm\)0.03 \\
    $\phi$ & 2.39 & 1.69 & 1.67 & 2.96 & 2.25\(\pm\)0.07 & 2.35 & 1.89 & 1.70 & 2.97 & 2.23\(\pm\)0.07 & 2.37 & 2.26 & 2.00 & 3.43 & 2.51\(\pm\)0.11 \\
    \hline
    $swf$ & 1.82 & 0.93 & 0.89 & 3.00 & \textbf{1.64\(\pm\)0.14} & 2.04 & 1.51 & 0.98 & 2.78 & \textbf{1.78\(\pm\)0.09} & 1.99 & 1.78 & 1.15 & 3.23 & \textbf{2.06\(\pm\)0.15} \\
    \hline
  \end{tabular}
  \caption{Aggregate Results per experimental design. For the full RL, \(\gamma=0.99\). Standard errors are reported only for averages to reduce clutter.}
  \label{table:aggregate_results}
\end{table*}
The contextual bandit SP approaches the analytical optimal solution for self-regulating firms. This formulation, while resulting in a more adaptive policy framework, requires more granular environment data in training, which may be difficult to obtain in the real world. Finally, the full RL formulation yields the best results in terms of welfare, achieving improvements relative to the baseline optimum on average. However, it intervenes at higher frequencies, re-defining policy mechanisms at each timestep. In addition to requiring granular training data, this may also reduce its range of applications to highly dynamic or volatile markets with shorter cycles. Globally, our results reveal that it is possible for a dynamic SP to incentive fairness in markets while retaining or even improving economic productivity.
\subsection{RQ2: Impact on individual firms}
We continue with a discussion on the individual impact to welfare bore by firms.~\autoref{tab:welfare_changes} denotes the per-firm welfare comparisons between the analytical baselines in the fairness-agnostic and fairness-aware cases, and firms subjected to the RL SP's learned incentive mechanisms. We note that, while most firms experience improvements in individual welfare, firm \textbf{C}'s is significantly reduced. This is likely due to the almost parallel nature of the demand curves of the two groups, and the fact that the fairer price is higher than the profit-maximizing one.
\begin{table}[h!]
\centering
\renewcommand{\arraystretch}{1.3}
\begin{tabular}{lccccr}
\hline
\textbf{Firm} & \textbf{A} & \textbf{B} & \textbf{C} & \textbf{D} & $\overline{\texttt{swf}}$ \\
\hline
\(\texttt{swf}_{agnostic}\) & 0.21 & 0.54 & 1.47 & 2.61 & 1.21 \\
\hline
\(\texttt{swf}_{aware}\) & 1.93 & 0.77 & 1.66 & 2.92 & 1.82 \\
\(\texttt{swf}_{SP}\) & 1.99 & 1.78 & 1.15 & 3.23 & 2.06 \\
\hline
\textbf{\(\texttt{swf}\%\Delta^{aware}_{SP}\)} & 3.11\% & 131.17\% & -30.72\% & 10.62\% & $\mathbf{13.19}\%$ \\
\hline
\end{tabular}
\caption{Summary of the welfare effects of the social planner's learned taxation and redistribution scheme compared to analytical baselines. The bottom row denotes the percentage change in welfare between the fairness-aware baseline and the social planner's generated \textit{fairness tax}.}
\label{tab:welfare_changes}
\end{table}
Nonetheless, the policy mechanism applied by the social planner RL agent significantly improves welfare outcomes compared to the fairness-agnostic case, and even manages to surpass those in the (unrealistic) case where firms self-regulate for fairness, with a 13.19\% improvement in global welfare. We note further that while global welfare improves under this setting, it is impossible for welfare to improve for every individual firm. Firm~\textbf{C}'s welfare loss indicates that there is indeed no such thing as a free lunch.

\subsection{RQ3: Deployment and performance over time}
\begin{figure*}[ht!]
    \centering
    % Adjust the widths of minipages if needed
    \begin{subfigure}[b]{0.45\linewidth}
        \includegraphics[width=\linewidth]{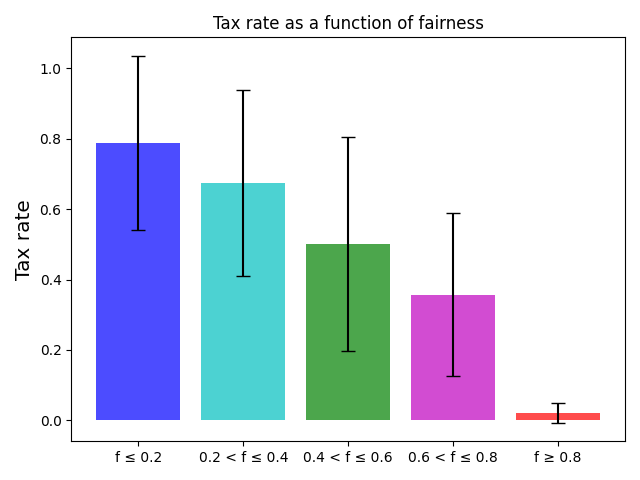}
        % \caption{Learned tax schedule for the full RL agent}
    \end{subfigure}
    % \hfill% Adds horizontal space between the minipages
    \begin{subfigure}[b]{0.545\linewidth}
        \includegraphics[width=\linewidth]{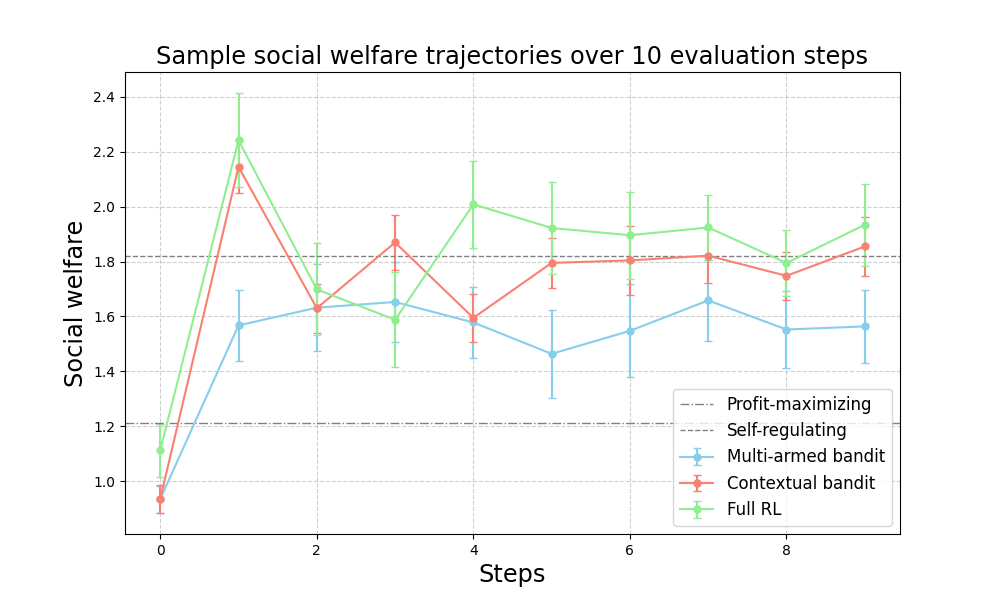}
        % \caption{Fairness trajectory over 10 evaluation steps.}
        % \label{fig:trajectory}
    \end{subfigure}
    \caption{\textit{Left}: Tax actions taken by the RL social planner. Reported policy mechanisms record SP actions averaged over 20 seeds. \textit{Right}: Social welfare trajectories during evaluation for the SP's learned policy frameworks from multi-armed bandit, contextual bandit, and RL formulations.}
    \label{fig:tax_and_trajectory}
\end{figure*}

%\subsection{Learned Incentive Mechanism}
As a high social impact application, it is necessary to study the usability and long-term impact of our framework. This is particularly relevant in the case that a similar solution is used to inform and assist policy makers in mechanism design. We note in~\autoref{fig:tax_and_trajectory} the emergence of a clear pattern between fairness and taxation, in line to a large degree with human intuition: firms demonstrating fairer behaviours are incentivized by lower tax rates, and this trend persists for each problem formulation (\autoref{fair_replay_buffer_appendix}).
In addition, each formulation of the SP learns to give the majority of the subsidy budget to the underrepresented consumer group (noted in~\autoref{table:aggregate_results}). For example, a proportion \(S=0.66\) means that for every unit of tax budget the social planner collects, it redistributes \(0.66\) units to \textit{group 1}, and \(0.34\) units to \textit{group 2}, thereby reducing their effective price and increasing their market participation. In our evaluation of welfare over time, the full RL SP outperforms both multi-armed and contextual bandit SPs in most timesteps, as shown in~\autoref{fig:tax_and_trajectory}.

\section{Limitations and Social Impact}
There exist opposing schools of thought regarding fiscal policy and policy intervention in general. While one would advocate for intervention for the sake of equality or social progressivity, another would claim that any form of intervention would disrupt the economic processes by which consumption and production are governed, and that any deviation from these processes is inefficient. In our simplified setting, we build an economic subsystem to illustrate the tradeoffs between policy intervention and its absence. Profit-maximizing firms are not likely to be intrinsically motivated to behave fairly with respect to their buyer distributions, especially when said behaviour entails a sacrifice in profitability. We illustrate this in~\autoref{table:baseline_multifirm_profit}, where profit-maximizing prices may have unforeseeable fairness outcomes dependent on distributions in demand. To mitigate this, we introduce a third-party social planner whose objective is to generate incentives which nudge firms into considering social welfare. Our experiments ultimately show that net improvements in social welfare are achievable with tailored taxation and redistribution schemes learned through a range of RL methods, each with their own applications. We note further that the adjusted prices paid by consumers (post-subsidy) are effective at increasing market participation.

\par Given that taxation and subsidy have important social implications, deploying AI-based strategies might not be easily accepted by decision-makers or the general public due to interpretability and accountability concerns. While we were able to generate seemingly interpretable tax schedules using RL, the chosen tax brackets themselves might not be easily accepted or interpreted, especially since they are linked with a penalty. Furthermore, while the resulting pattern of tax schedules is consistent across random seeds, the values for each tax bracket varies. Deployment of such a solution therefore requires additional considerations for robustness against distribution shifts and possible poisoning attacks. Additionally, dynamic pricing is becoming more algorithmically driven, thus requiring more studies into how these algorithms might adapt in response to our incentive strategies. 

\par Our research has several simplifying assumptions, making it difficult to utilize this framework in real-world scenarios.  For instance, considering only demand fairness might lead to counter-intuitive positive benchmarks, as seen in our examples where analytically-optimal prices are, in some cases, higher than their fairness-agnostic counterpart. Further, it is difficult to conceive of a mechanism for assessing fairness on a per-firm basis in the real world. This would presumably necessitate policy makers to access broader market data, as well as consumer-level data from individual firms, from which they may compute distributions which would be necessary for a fairness-based tax to be enforced. Additionally, our experiments consider a single-product market with no competitive dynamics between firms. While our setting illustrates a real concern for the disparate impacts of dynamic pricing on consumer distributions, introducing consumer choice would increase realism, but would perhaps make achieving improvements in welfare more challenging. Lastly, we focused on demand fairness given how it reflects accessibility to essential goods and services in critical domains such as healthcare, but further research is necessary to explore how our framework can be applied to alternative fairness notions such as price and consumer surplus fairness, which expand to other applications such as loan approval. 

\par In a profit-driven market, our work shows that merely relying on self-regulation is not sufficient, thus requiring interventions from policy makers to ensure equity and equal access to multiple goods and services. While we do not present our framework as the ultimate solution, we believe it can guide future research and inspire exploration at the intersection of AI and society, particularly in the design of dynamic incentive mechanisms, especially in applications where dynamic pricing has high social and economic impacts.

\section{Conclusion}
When used for fairness-agnostic profit-maximization, dynamic pricing can have harmful consequences with respect to equal access. We demonstrate that, by leveraging the policy tools of taxation and subsidy, a dynamic social planner may create global efficiencies which would have otherwise been unattainable, even with locally optimal firm behaviour by self-regulation. This not only reveals that a dynamic third party may be trained to improve fairness outcomes in a broad sense, but may also generate policies which are tailored to specific markets and consumer dynamics, and that these policies may be deployed in ways that improve upon free-market outcomes. While we believe these findings to be relevant for monopolistic markets or geographically-delimited markets with little potential for consumer displacement, we see great value in extending this work to competitive markets, where consumers have more freedom in choosing among different products. Additionally, exploring more realistic and challenging environments with imbalanced consumer distributions is necessary to further understand and analyze the financial and social implications of such frameworks. Finally, extending the framework to use various fairness definitions would be needed to ensure its compatibility with other application domains. 

\begin{acks}
\label{sec:ack}
Funding support for project activities has been partially provided by Canada CIFAR AI Chair, Google award, MEI award, and the Natural Sciences and Engineering Research Council of Canada (NSERC). We also express our gratitude to Compute Canada for their support in providing facilities for our evaluations. Finally, we thank Nobuhiro Kiyotaki for the valuable insight he provided during the early ideation stages of this research.
\end{acks}
%%%%%%%%%%%%%%%%%%%%%%%%%%%%%%%%%%%%%%%%%%%%%%%%%%%%%%%%%%%%%%%%
%% Bibliography
%%%%%%%%%%%%%%%%%%%%%%%%%%%%%%%%%%%%%%%%%%%%%%%%%%%%%%%%%%%%%%%%
\bibliographystyle{ACM-Reference-Format}
\bibliography{sample-base}

% %%%%%%%%%%%%%%%%%%%%%%%%%%%%%%%%%%%%%%%%%%%%%%%%%%%%%%%%%%%%%%%%
% %% Appendices
% %%%%%%%%%%%%%%%%%%%%%%%%%%%%%%%%%%%%%%%%%%%%%%%%%%%%%%%%%%%%%%%%
\newpage
\appendix
\onecolumn
\section{Demand Distribution Parameters}
\label{appendix:consumer_distribution_params}
In our experiments, we designed four firms A, B, C and D who each address a customer base whose demand distributions depend on parameters sampled in the table below.

\begin{center}
\begin{threeparttable}
    \begin{tabular}{c*{8}{c}}
        \toprule
        & \multicolumn{2}{c}{Firm A} & \multicolumn{2}{c}{Firm B} & \multicolumn{2}{c}{Firm C} & \multicolumn{2}{c}{Firm D} \\
        \cmidrule(lr){2-3} \cmidrule(lr){4-5} \cmidrule(lr){6-7} \cmidrule(lr){8-9}
        Params & $g1$ & $g2$ & $g1$ & $g2$ & $g1$ & $g2$ & $g1$ & $g2$ \\
        \midrule
        \textit{w} & -1.926 & -2.369 & -1.9 & -0.695 & -0.340 & -0.600 & -2.369 & -1.1526 \\
        \textit{b} & 6.4757 & 15.7900 & 5.4757 & 5.229 & 0.9195 & 4.4757 & 10.2290 & 8.4757 \\
        \bottomrule
    \end{tabular}
    \begin{tablenotes}
        \small
        \item[*] In stochastic experiments, values are sampled normally around table values with a standard deviation of \(\sigma = 0.1\) every 50,000 steps, and \(\sigma = 0.0125\) in intermediate steps.
    \end{tablenotes}
\end{threeparttable}
\end{center}

\section{Additional Experiments}
\label{appendix:additional_experiments}

In addition to the experimental results reported in the main section, we also ran experiments over a variety of reward functions without the use of a subsidy. For instance, including gross profit (pre-tax) into the social planner's reward function made it less concerned for the after-tax income of firms, and more with the overall size of the economy measured by the sum of net-profit and a tax budget. This meant that firms were taxed at very high rates, and while fairness was improved, the profit values reported do not reflect the welfare of individual firms. 
\begin{table*}[h!]
  \centering
  \label{table:all_results}
  \begin{tabular}{c|c c c c c|c c c c c|c c c c c}
    \hline
    & \multicolumn{5}{c|}{Multi-armed bandit} & \multicolumn{5}{c|}{Contextual bandit} & \multicolumn{5}{c}{Full RL} \\
    \hline
    Firms & A & B & C & D & \textbf{Avg} & A & B & C & D & \textbf{Avg} & A & B & C & D & \textbf{Avg} \\
    \hline
    $f$ & 0.76 & 0.32 & 0.57 & 0.76 & 0.60 & 0.79 & 0.33 & 0.56 & 0.81 & 0.62 & 0.76 & 0.46 & 0.62 & 0.83 & 0.67 \\
    $\phi$ & 2.39 & 2.16 & 2.71 & 3.22 & 2.62 & 2.37 & 2.16 & 2.69 & 3.33 & 2.63 & 2.39 & 1.97 & 2.52 & 3.23 & 2.53 \\
    \hline
    $swf$ & 1.82 & 0.69 & 1.54 & 2.44 & \textbf{1.57} & 1.87 & 0.71 & 1.51 & 2.69 & \textbf{1.63} & 1.81 & 0.91 & 1.56 & 2.68 & \textbf{1.70} \\
    \hline
  \end{tabular}
  \caption{Aggregate Results w/ Gross Profit}
\end{table*}

To evaluate firm welfare in the absence of subsidy, the social planner considers net profit in its reward. In this setting, it faces the tradeoff between welfare and taxation, but the tax budget is not reinjected into the economy, making it difficult to draw comparisons with the analytical baselines, where firms are self-regulating and do not pay taxes. Nonetheless, fairness improvements are still achievable.
\begin{table*}[h!]
  \centering
  \label{table:all_results}
  \begin{tabular}{c|c c c c c|c c c c c|c c c c c}
    \hline
    & \multicolumn{5}{c|}{Multi-armed bandit} & \multicolumn{5}{c|}{Contextual bandit} & \multicolumn{5}{c}{Full RL} \\
    \hline
    Firms & A & B & C & D & \textbf{Avg} & A & B & C & D & \textbf{Avg} & A & B & C & D & \textbf{Avg} \\
    \hline
    $f$ & 0.79 & 0.33 & 0.58 & 0.69 & 0.60 & 0.62 & 0.36 & 0.57 & 0.78 & 0.58 & 0.73 & 0.27 & 0.55 & 0.77 & 0.58 \\
    $\phi$ & 1.98 & 1.19 & 1.35 & 2.19 & 1.43 & 2.07 & 1.42 & 2.44 & 2.91 & 2.21 & 2.30 & 1.52 & 2.26 & 2.89 & 2.24 \\
    \hline
    $swf$ & 1.56 & 0.40 & 0.79 & 1.52 & \textbf{0.86} & 1.28 & 0.51 & 1.39 & 2.27 & \textbf{1.28} & 1.68 & 0.41 & 1.25 & 2.23 & \textbf{1.30} \\
    \hline
  \end{tabular}
  \caption{Aggregate Results w/ Net Profit.}
\end{table*}

\newpage
\section{Fair Replay Buffer}
\label{fair_replay_buffer_appendix}
Here, we include details regarding the~\texttt{FairReplayBuffer}, including the algorithm along with ablations highlighting the differences between the tax schedules learned by the SP agent when using FIFO, and~\texttt{FairReplayBuffer}.
\subsection{Algorithm}
\begin{algorithm}[ht]
\caption{\texttt{FairReplayBuffer}}
\label{alg:fair_buffer}

\DontPrintSemicolon
\KwIn{buffer capacity $|\mathcal{D}|$, $obs$, $action$, $reward$, $done$, $\mathit{infos}$, brackets $\mathcal{I}$, batch size $|\mathcal{B}|$}
\KwOut{A replay buffer ensuring uniform distribution of experiences across the fairness-profit space}

Buffer initialization, $\mathcal{D} \gets [\;]$\;
%Initialize $\mathit{buffer}$ as an empty list\;
% $\mathcal{I} \gets \emptyset$\;
%Initialize $\mathit{brackets}$ as an empty set\;
Storage per bracket initialization, $\mathcal{S} \gets \{\}$\;
%Initialize $\mathit{storage\_per\_bracket}$ as an empty dictionary\;
% Counter initialization, $c \gets \{\}$\;
%Initialize $\mathit{count\_per\_bracket}$ as an empty dictionary\;

% \tcc{Initialization for each bracket}
% \For{$i$ in $\mathcal{I}$}{
%     %$\mathit{count\_per\_bracket}[bracket] \gets$ empty list\;
% % $c[i] \gets []$\;
%     $c[i] \gets 0$\;
% }

\tcc{Procedure to add experiences}
\SetKwFunction{FAdd}{ADD}
\SetKwProg{Fn}{Procedure}{:}{}
\Fn{\FAdd{$\mathit{obs, action, reward, done, infos}$}}{
    Determine $i$ based on $\mathit{obs}$\;
    % \If{$c[i] < |\mathcal{D}| / |\mathcal{I}|$}{
    Append$(\mathit{obs, action, reward, done, infos})$ to $\mathcal{D}$\;
    Append index of new experience to $\mathcal{S}[i]$\;
    % $c[i] \gets c[i] + 1$\;
    % }
}

\tcc{Procedure to sample experiences}
\SetKwFunction{FSample}{SAMPLE}
\Fn{\FSample{$|\mathcal{B}|$}}{
    Sampled batch initialization, $\mathcal{B}  \gets [\;]$\;
    $|b| \gets |\mathcal{B}| / |\mathcal{I}|$\;
    \For{$i$ in $\mathcal{I}$}{
        $b \gets$ Sample $(o, a, r, d, \mathit{info})$ from $\mathcal{S}[i]$\ ; \\
        Add $b$ to $\mathcal{B}$\;
    }
    Shuffle $\mathcal{B}$\;
    \KwRet $\mathcal{B}$\;
}
\end{algorithm}

\subsection{Ablations}
\label{ablation}
We ablate the~\texttt{FairReplayBuffer} by comparing it with trials run with a first-in-first-out (FIFO) buffer. The intention behind designing this replay buffer was to generate a tax schedule with intuitive patterns, achieved by ensuring that the learning agent retains information from less frequently-observed fairness brackets.
\begin{figure}[htbp]
    \centering
    \begin{subfigure}{0.40\textwidth}
        \includegraphics[width=\linewidth]{tax_schedules/2d_ub_fairness_profit-TRIAL-RL-sub.png}
        \caption{RL,~\texttt{FairReplayBuffer}}
        \label{fig:rl_frb_schedule}
    \end{subfigure}\hfill
    \begin{subfigure}{0.4\textwidth}
        \includegraphics[width=\linewidth]{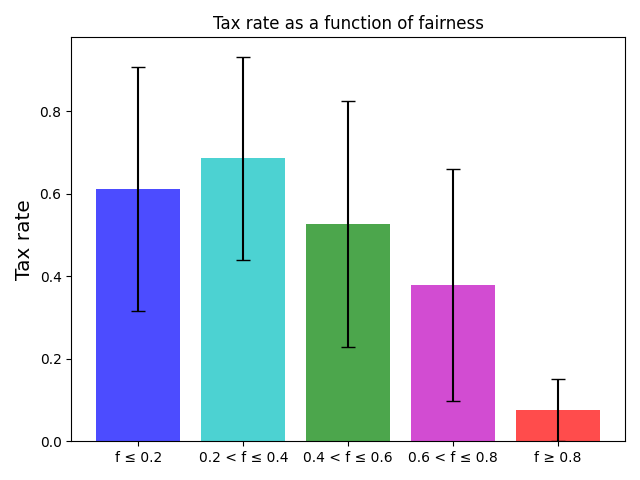}
        \caption{RL,~\texttt{FIFO}}
        \label{fig:rl_fifo_schedule}
    \end{subfigure}
    \caption{\texttt{FairReplayBuffer} vs. FIFO for the RL setting}
    \label{fig:rl_frb_fifo}
\end{figure}

\begin{figure}[htbp]
    \centering
    \begin{subfigure}{0.40\textwidth}
        \includegraphics[width=\linewidth]{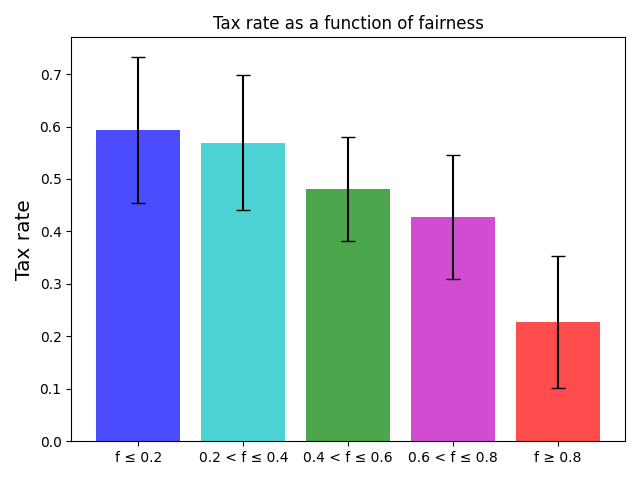}
        \caption{C-MAB,~\texttt{FairReplayBuffer}}
        \label{fig:cmab_frb_schedule}
    \end{subfigure}\hfill
    \begin{subfigure}{0.4\textwidth}
        \includegraphics[width=\linewidth]{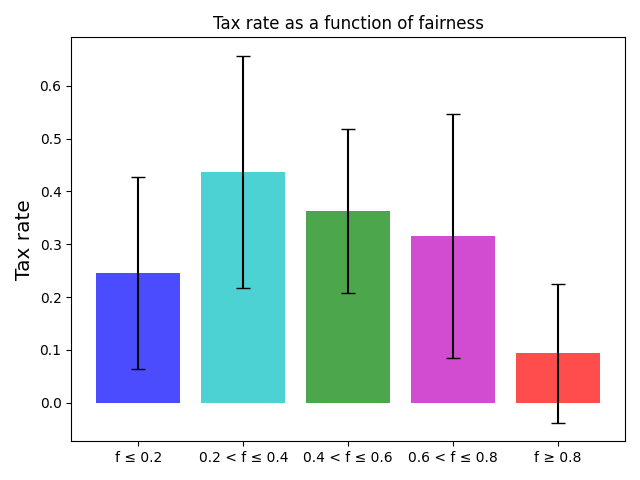}
        \caption{C-MAB,~\texttt{FIFO}}
        \label{fig:cmab_fifo_schedule}
    \end{subfigure}
    \caption{\texttt{FairReplayBuffer} vs. FIFO for the C-MAB setting}
    \label{fig:cmab_frb_fifo}
\end{figure}

\begin{figure}[htbp]
    \centering
    \begin{subfigure}{0.40\textwidth}
        \includegraphics[width=\linewidth]{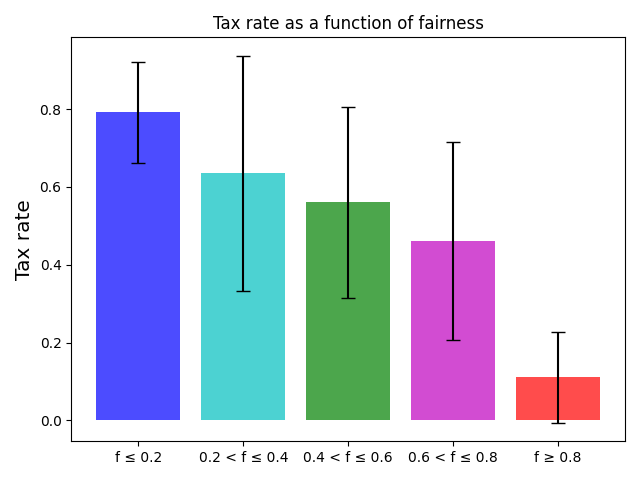}
        \caption{MAB,~\texttt{FairReplayBuffer}}
        \label{fig:mab_frb_schedule}
    \end{subfigure}\hfill
    \begin{subfigure}{0.4\textwidth}
        \includegraphics[width=\linewidth]{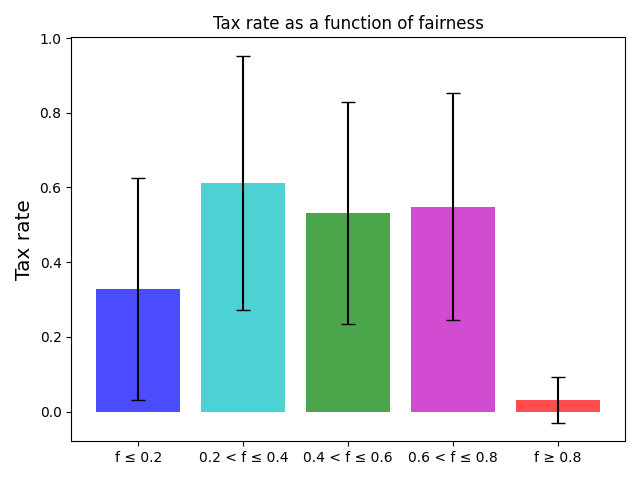}
        \caption{MAB,~\texttt{FIFO}}
        \label{fig:mab_fifo_schedule}
    \end{subfigure}
    \caption{\texttt{FairReplayBuffer} vs. FIFO for the MAB setting}
    \label{fig:mab_frb_fifo}
\end{figure}

\end{document}